\documentclass[letterpaper, 10 pt, conference]{ieeeconf}  

\IEEEoverridecommandlockouts                              

\usepackage{amsmath,amsfonts,amssymb,amsthm}
\usepackage{array}
\usepackage[caption=false,font=normalsize,labelfont=sf,textfont=sf]{subfig}
\usepackage{textcomp}
\usepackage{stfloats}
\usepackage{url}
\usepackage{verbatim}
\usepackage{graphicx}
\usepackage{cite}
\usepackage{xcolor}
\usepackage{threeparttable}
\usepackage{booktabs}

\usepackage[ruled,vlined,linesnumbered]{algorithm2e}

\usepackage{array,booktabs,tabularx}
\newcolumntype{Y}{>{\centering\arraybackslash}X}

\theoremstyle{plain}

\theoremstyle{definition}
\newtheorem{problem}{Problem}

\theoremstyle{remark}

\title{\LARGE \bf
Passage-Focused Soft Coordination for Hierarchical Multi-Robot Navigation in Bottleneck Environments
}

\author{Haoze Dong$^{\dagger, 1}$, Stepan Dergachev$^{\dagger, 2}$,
Meng Guo$^{1}$, Konstantin Yakovlev$^{3}$, and Zhongkui Li$^{1}$%
\thanks{$^{\dagger}$Haoze Dong and Stepan Dergachev contributed equally
to this work.}%
\thanks{$^{1}$Haoze Dong, Meng Guo, and Zhongkui Li are with the
School of Advanced Manufacturing and Robotics, Peking University,
Beijing, China.}%
\thanks{$^{2}$Stepan Dergachev is with the National Research University
Higher School of Economics, Moscow, Russia.}%
\thanks{$^{3}$Konstantin Yakovlev is with Saint-Petersburg University,
Saint Petersburg, Russia.}%
\thanks{Stepan Dergachev ({\tt sadergachev@hse.ru}) and Zhongkui Li ({\tt zhongkli@pku.edu.cn}) are the corresponding authors.}%
}

\begin{document}

\maketitle
\thispagestyle{empty}
\pagestyle{empty}




\begin{abstract}

In multi-robot navigation, rigid global schedules are sensitive to execution deviations, while local methods lack long-horizon coordination. We propose a passage-focused framework combining soft global coordination with flexible local execution. A passage-induced graph of convex sets supports priority-ordered planning with two levels of soft coordination. Passage-aware route selection accounts for passage usage and traversal directions, while route-restricted continuous space-time planning penalizes temporal conflicts. The resulting references guide a model predictive path integral controller that avoids collisions online. Adaptive temporal and geometric tracking accommodates execution deviations, while selective replanning maintains coordination without rigid trajectory reservations. Across 1,200 instances in 4 different environments with 5–30 robots, the framework achieves a near-perfect success rate and outperforms evaluated baselines in makespan and flowtime. Hardware experiments with 8 robots demonstrate successful completion of all trials without collisions. These results highlight the effectiveness of passage-focused coordination in balancing long-horizon cooperation with flexible execution.

\end{abstract}

\section{Introduction}

Multi-robot navigation in cluttered environments is particularly challenging at narrow passages, where robots sharing limited space must coordinate both their route choices and traversal times.
Global multi-agent planning can organize these decisions through routes and schedules, but approaches that tightly prescribe collision-free paths or trajectories may require revisions when tracking errors, delays, or local avoidance maneuvers disrupt nominal execution~\cite{atzmon2020robust,chen2021scalable}.
In contrast, online trajectory optimization and collision-avoidance methods can adapt to nearby robots and obstacles~\cite{dergachev2025decentralized,chen2023multi}, yet local reactions alone may fail to anticipate conflicting passage demands, resulting in congestion or deadlock~\cite{dong2025homotopy}.
The key challenge is thus to coordinate bottleneck traversal over long horizons without rigidly constraining how robots respond to local interactions during execution.

\subsection{Related Work}

Multi-agent path finding (MAPF) coordinates robots through conflict-free paths on discrete graphs~\cite{stern2019multi}.
CBS and EECBS provide optimal or bounded-suboptimal solutions~\cite{sharon2015conflict,li2021eecbs}, while priority-based approaches emphasize computational efficiency~\cite{silver2005cooperative,okumura2023lacam}.
Although classical MAPF relies on discrete space-time schedules, robust and continuous-time extensions accommodate delays and nonuniform action durations~\cite{atzmon2020robust,andreychuk2022multi}, while hybrid approaches use MAPF guidance for continuous trajectory optimization~\cite{park2025decentralized}.
Nevertheless, coordination remains largely tied to specific discrete paths and space-time occupancy commitments, limiting spatial and temporal flexibility for local adaptation during execution.

Kinodynamic multi-robot planners bridge discrete coordination and continuous execution by incorporating motion constraints into global planning.
Conflict-based methods account for robot dynamics~\cite{kottinger2022conflict}, while discontinuity-bounded search improves planning scalability~\cite{moldagalieva2024dbcbs,moldagalieva2026dblacam}.
Space-time graphs of convex sets (ST-GCS) enable continuous trajectory planning with flexible timing and velocity bounds, using space-time reservations for inter-robot collision avoidance~\cite{tang2025stgcs}; search-based extensions further improve efficiency~\cite{tang2026search}.
However, these reservation-based approaches still assign primary collision-avoidance responsibility to global planning, making coordination sensitive to execution deviations.


\begin{figure}[t]
    \centering
    \includegraphics[width=\linewidth]{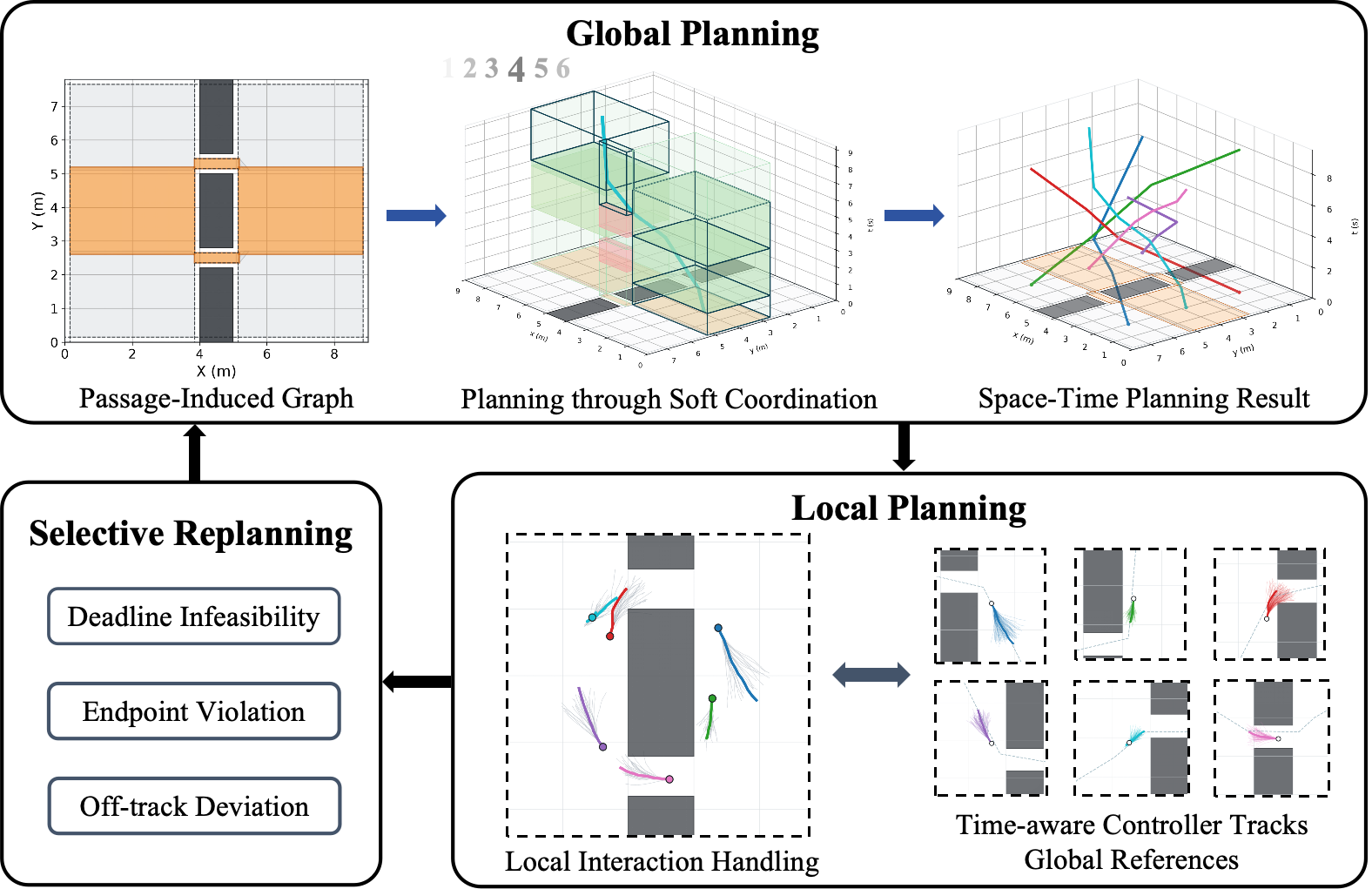}
    \caption{
    Overview of the proposed hierarchical multi-robot navigation framework. The global planner coordinates passage selection and traversal timing through soft space–time conflict penalties, while the local time-aware MPPI–based controller handles collision avoidance and execution deviations, with selective replanning maintaining coordination.
    }
    \label{fig:framework}
    \vspace{-0.4cm}
\end{figure}

Online local methods adapt robot motion to execution deviations and nearby interactions. Reactive methods such as ORCA and BVC use local state information for decentralized collision avoidance~\cite{berg2011reciprocal,zhou2017fast}, while distributed trajectory optimization exchanges trajectories for longer-horizon coordination~\cite{tordesillas2021mader,chen2023multi,le2024multi}. Model predictive path integral (MPPI) accommodates nonlinear dynamics but uses cost penalties for collision avoidance. MPPI-ORCA~\cite{dergachev2024model} incorporates hard constraints for decentralized collision avoidance without trajectory exchange, with~\cite{dergachev2025decentralized} addressing sensing and execution uncertainty.

However, purely local methods may fail to anticipate competing passage demands, causing congestion or deadlock.
Hierarchical approaches combine passage-aware global guidance with local optimization and replanning~\cite{dong2025homotopy}, but estimating passage occupancy from average speeds limits proactive adjustment of traversal timing to mitigate bottleneck conflicts.

Together, these limitations motivate a different allocation of planning responsibilities, where global coordination anticipates bottleneck conflicts without prescribing collision-avoidance maneuvers during local execution.

\subsection{Our Method}

As shown in Fig.~\ref{fig:framework}, 
we develop a hierarchical multi-robot navigation framework that assigns complementary roles to global and local planning: the global layer softly coordinates bottleneck traversal, while the local layer adapts motions and resolves collisions online.
At the global level, priority-ordered planning on a passage-induced graph of convex sets first selects spatial routes based on passage usage and traversal directions, then optimizes traversal timing within the selected region tubes using soft space-time conflict penalties.
Accepted references update coordination information, guiding subsequent robots without imposing hard inter-robot trajectory reservations.
At the local level, a time-aware MPPI-based controller handles inter-robot and obstacle avoidance while tracking globally coordinated references.
Adaptive temporal and spatial tracking allows robots to accommodate execution deviations without rigidly following nominal schedules.
To maintain coordination during execution, selective global replanning updates inconsistent references while preserving the plans of unaffected robots.

The main contributions of this work are threefold:
(I) We develop a passage-focused global planner that couples passage-aware spatial route selection with route-restricted continuous space-time planning through two complementary levels of soft inter-robot coordination.
(II) We extend a multi-agent MPPI-ORCA controller with explicit static-obstacle avoidance and adaptive temporal and geometric reference tracking, enabling flexible execution of globally coordinated references while handling online robot interactions.
(III) We establish a hierarchical coordination framework that separates long-horizon passage coordination from short-horizon collision avoidance, using time-parameterized references as soft guidance rather than strict execution commitments. Selective global replanning maintains coordination consistency as actual execution deviates from the planned references.
\section{Problem Formulation}

Consider a set of $N$ homogeneous disk robots $\mathcal A \triangleq \{1,\ldots,N\}$ of radius $r$ navigating in a shared workspace $\mathcal W\subset\mathbb R^2$ containing static convex polygonal obstacles $\{\mathcal O^m\}$. Let $\widetilde{\mathcal O}^m$ denote obstacle $\mathcal O^m$ inflated by $r$, and define the collision-free workspace as $\mathcal F \triangleq \mathcal W\setminus\bigcup_m\widetilde{\mathcal O}^m$. Each robot $i\in\mathcal A$ starts from a state $\mathbf s_i$ whose position lies in $\mathcal F$ and must reach a goal position $\mathbf g_i\in\mathcal F$.
For closed-loop execution, let $k$ denote the discrete control step.
Let $\mathbf x_{i,k}$ and $\mathbf u_{i,k}$ denote the state and control input of robot $i$ at time step $k$, with $\mathbf p_{i,k}$ denoting the position component of its state. The robot obeys the discrete-time control-affine dynamics
\begin{equation}
    \mathbf{x}_{i,k+1}=F_i(\mathbf{x}_{i,k}) + G_i(\mathbf{x}_{i,k})\mathbf{u}_{i,k},
\qquad
\mathbf u_{i,k}\in\mathcal U_i,
\label{eq:motion_model}
\end{equation}
where $F_i$ describes the uncontrolled state evolution, $G_i$ maps control inputs to state changes, and the bounded set $\mathcal U_i$ specifies the admissible controls.
A global reference is a time-parameterized path $\pi_i:[0,t_i^{f}]\rightarrow\mathcal F$ satisfying
\begin{equation}
\pi_i(0)=\mathbf p_{i,0},\qquad
\pi_i(t_i^{f})=\mathbf g_i,\qquad
\|\dot{\pi}_i(t)\|_2\leq\bar v,
\end{equation}
where $t_i^{f}$ is its duration and $\bar v$ is the speed bound used for global planning. 
An executed trajectory is a sequence of states $\{\mathbf x_{i,k}\}_{k=0}^{K_i}$, where $K_i$ is its duration in discrete time steps.  A collection of trajectories is valid if it satisfies the dynamics and control limits above and
\begin{equation}
\begin{gathered}
\mathbf{x}_{i,0}=\mathbf{s}_i,\quad
\mathbf{p}_{i,K_i}\in\mathcal{B}_\varepsilon(\mathbf{g}_i),\quad
\mathbf{p}_{i,k}\in\mathcal{F},\\
\|\mathbf{p}_{i,k}-\mathbf{p}_{j,k}\|_2\geq 2r,
\quad \forall i\neq j ,\  \forall k
\end{gathered}
\label{eq:problem_formulation}
\end{equation}
where $\mathcal B_\varepsilon(\mathbf g_i)$ denotes the closed Euclidean ball of radius $\varepsilon$ centered at $\mathbf g_i$.

\begin{problem}
Given the workspace $\mathcal W$, obstacles $\{\mathcal O^m\}$, initial states and goals $\{(\mathbf s_i,\mathbf g_i)\}_{i\in\mathcal A}$, and robot dynamics and control limits $(F_i,G_i,\mathcal U_i)$, determine global references $\{\pi_i\}_{i\in\mathcal A}$ and closed-loop control inputs $\{\mathbf u_{i,k}\}_{k=0}^{K_i-1}$ for each robot that generate a valid collection of trajectories satisfying~\eqref{eq:problem_formulation}.
\end{problem}

\section{Method}


Our framework combines long-horizon passage-focused global planning, short-horizon MPPI-based local control, and selective replanning. The global planner provides soft time-parameterized references, while the local controller resolves collisions and accommodates execution deviations.

\subsection{Passage-Focused Global Planning}


The global planner coordinates bottlenecks traversal through priority-ordered route selection and space-time optimization, using passage-usage records and space-time conflict indices to guide subsequent robots.

\subsubsection{Passage-Induced Graph of Convex Sets}
\label{subsec:passage_gcs}

\begin{figure}[t]
    \centering
    \includegraphics[width=\linewidth]{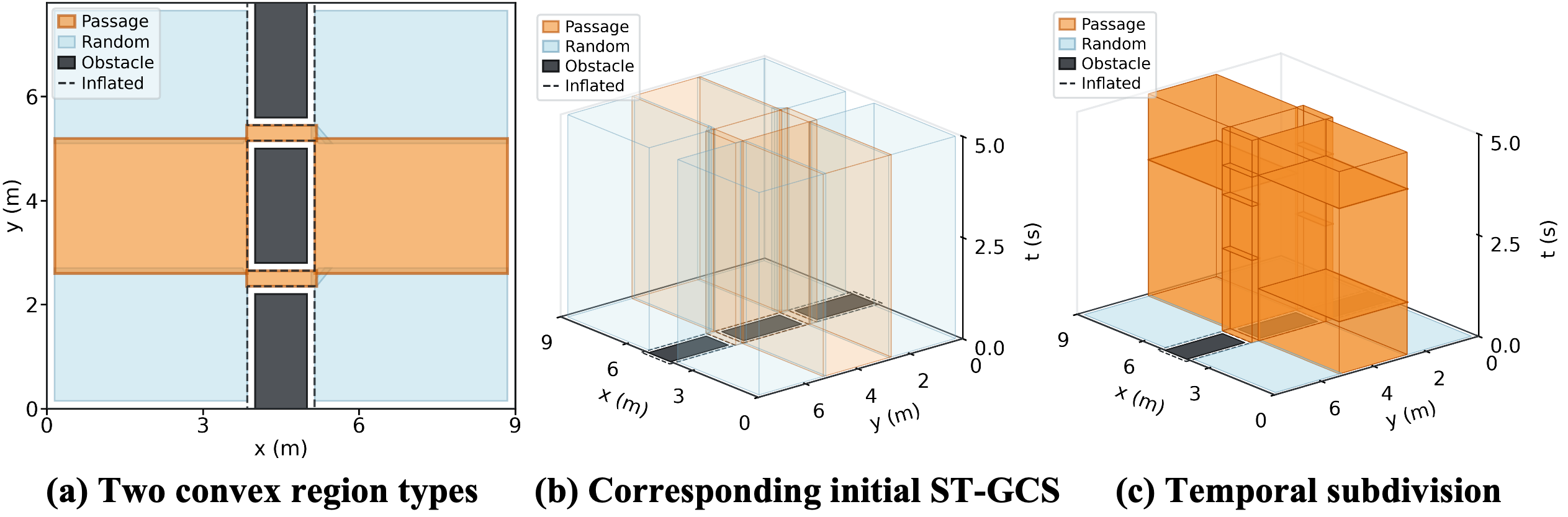}
    \caption{
    Construction of the passage-induced ST-GCS. (a) Spatial convex regions, including passage regions (orange) and randomly sampled regions (blue). (b) Extension of the spatial regions into the time dimension. (c) Temporal subdivision of passage-region space-time sets.
    }
    \label{fig:gcs_visualization}
    \vspace{-0.4cm}
\end{figure}


We represent narrow passages as convex regions within a spatial graph for route selection, then extend the regions along the time dimension to form an ST-GCS for space-time planning.
For each valid obstacle pair \((\widetilde{\mathcal{O}}^{a},\widetilde{\mathcal{O}}^{b})\), we apply the complete passage detection method of~\cite{dong2025homotopy} to obtain the shortest connecting segment:
$
(\mathbf{p}_{*}^{a},\mathbf{p}_{*}^{b})
=
\arg\min_{\mathbf{p}^{a}\in\widetilde{\mathcal{O}}^{a},
\,
\mathbf{p}^{b}\in\widetilde{\mathcal{O}}^{b}}
\left\|\mathbf{p}^{a}-\mathbf{p}^{b}\right\|_{2},
$
with passage width
$
w_{ab}
=
\left\|\mathbf{p}_{*}^{a}-\mathbf{p}_{*}^{b}\right\|_{2}.
$
Following~\cite{dong2025homotopy}, this segment is translated in both perpendicular directions until the passage-validity conditions cease to hold, yielding the two extreme cross-sections
$l(\mathbf{p}_{+}^{a},\mathbf{p}_{+}^{b})$ and
$l(\mathbf{p}_{-}^{a},\mathbf{p}_{-}^{b})$.
The complete passage is represented by
$
\mathcal{P}^{ab}
\triangleq
\left(
l(\mathbf{p}_{+}^{a},\mathbf{p}_{+}^{b}),
l(\mathbf{p}_{*}^{a},\mathbf{p}_{*}^{b}),
l(\mathbf{p}_{-}^{a},\mathbf{p}_{-}^{b})
\right).
$
Each complete passage is then represented by a convex region:
\begin{equation}
\mathcal{C}^{P}_{ab}
\triangleq
\bigcap_{j=1}^{4}
\left\{
\mathbf{x}\in\mathbb{R}^{2}
\,\middle|\,
\mathbf{a}_{ab,j}^{\top}\mathbf{x}\leq b_{ab,j}
\right\}.
\label{eq:passage_convex_region}
\end{equation}
Two half-spaces separate the region from the generating obstacles, while the remaining two are induced by the extreme passage cross-sections.  
Additional convex regions are generated from collision-free samples to cover the remaining free space (see Fig.~\ref{fig:gcs_visualization}(a)).
Let
$\{\mathcal{C}_{k}\}_{k\in\mathcal{V}}$
denote all resulting regions and let
$\mathcal{V}_{P}\subseteq\mathcal{V}$
denote the passage-region indices.
These regions form an undirected region graph
\begin{equation}
\mathcal{G}_{R}
\triangleq
(\mathcal{V},\mathcal{E}),
\qquad
(u,v)\in\mathcal{E}
\Longleftrightarrow
\mathcal{C}_{u}\cap\mathcal{C}_{v}\neq\emptyset.
\label{eq:region_graph}
\end{equation}
For each adjacent pair, a feasible point
$
\mathbf{q}_{uv}
\in
\mathcal{C}_{u}\cap\mathcal{C}_{v}
$
is selected as an inter-region portal. 

Following~\cite{tang2025stgcs}, we lift the spatial region graph into the time dimension. 
Let
$
\mathbf{z}
=
[\mathbf{p}^{\top},t]^{\top}
\in
\mathbb{R}^{3}.
$
The resulting ST-GCS is
\begin{equation}
\mathcal{G}_{ST}
\triangleq
\left(
\mathcal{V}_{ST},
\mathcal{E}_{ST},
\{\mathcal{X}_{v}\}_{v\in\mathcal{V}_{ST}}
\right),
\ 
\mathcal{X}_{v}
\subseteq
\mathcal{C}_{k(v)}
\times
[0,t_{\max}],
\label{eq:stgcs_definition}
\end{equation}
where \(k(v)\in\mathcal V\) denotes the parent spatial region. Initially, each spatial region is extruded over \([0,t_{\max}]\), while temporal subdivision may create multiple vertices sharing the same parent region (see Fig.~\ref{fig:gcs_visualization}(b)-(c)). 
Edges are defined by
$
(u,v)\in\mathcal{E}_{ST}
\Longleftrightarrow
\mathcal{X}_{u}\cap\mathcal{X}_{v}\neq\emptyset.
$
Vertices inherited from passage regions additionally retain their passage width and coordination metadata.

\subsubsection{Priority-Ordered Space-Time Global Planning}
\label{subsec:priority_st_global_planning}

Robots are planned in a fixed priority order, where \(j \prec i\) means that robot \(j\) has higher priority than robot \(i\). ST-GCS supports continuous space-time planning under velocity constraints~\cite{tang2025stgcs}, while search-based GCS methods interleave graph exploration with continuous optimization~\cite{natarajan2024implicit,tang2026search}. Since incorporating previously planned trajectories throughout the full space-time graph can increase search complexity, we adopt a two-stage strategy: each robot first selects a passage-aware spatial route and then optimizes its space-time path in the route-induced ST-GCS. 
After each robot is planned, its reference updates two coordination signals used for subsequent robots.










\begin{figure}[t]
    \centering
    \includegraphics[width=\linewidth]{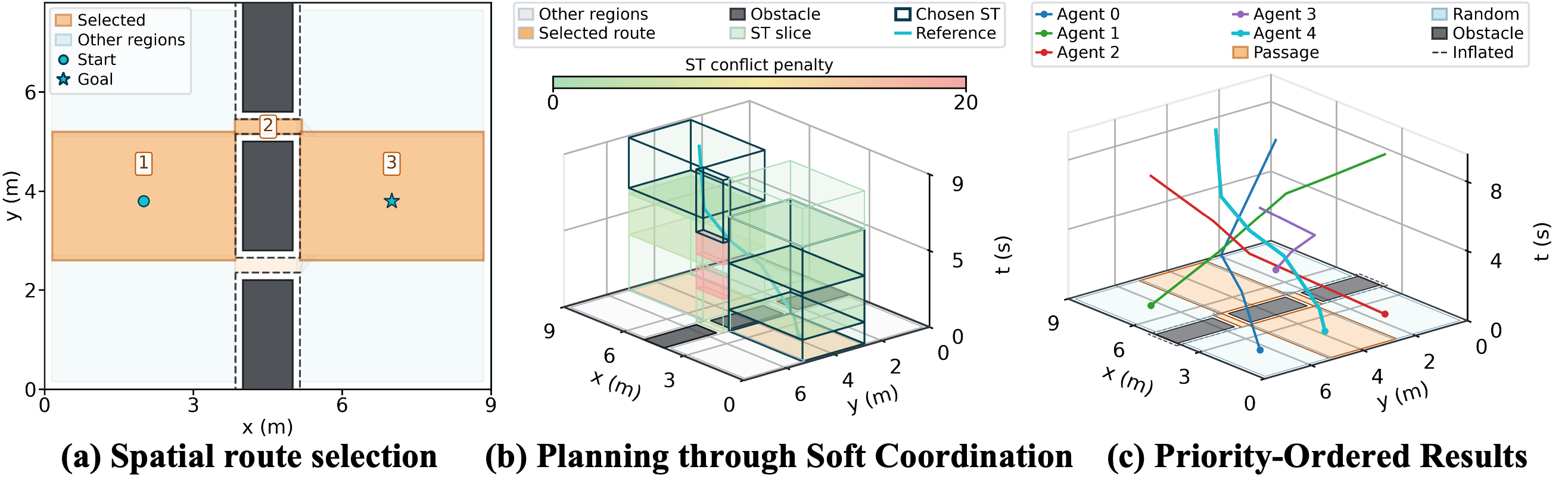}
    \caption{
    Illustration of passage-focused global planning. (a) Passage-aware spatial route selection. (b) Route-restricted ST-GCS with temporally subdivided regions shaded by soft conflict penalties, the selected space-time vertex sequence (dark outlines), and the resulting reference path (cyan). (c) Time-parameterized reference paths for five robots planned in priority order.
    }
    \label{fig:single-agent-planning}
    \vspace{-0.4cm}
\end{figure}

At planning time $\tau$, let $\mathbf{p}_i(\tau)$ denote robot $i$'s current position, with $\mathbf{p}_i(0)=\mathbf{p}_{i,0}$.
We augment the region graph $\mathcal{G}_{R}$ with query-specific source and goal connections to compatible regions. Since the geometric transition cost depends on consecutive portals, route search is performed over expanded states
$
q=(u,v),
$
where $u$ and $v$ denote the previous and current regions. A candidate route is represented as
$
\rho_i \triangleq (k_{i,1},\ldots,k_{i,L_i}).
$
Its cost is
\begin{equation}
J_i^R(\rho)
\triangleq
J^G(\rho)
+
J_i^P(\rho),
\label{eq:region_route_cost}
\end{equation}
where $J^G$ is the portal-polyline length from $\mathbf{p}_i(\tau)$ to $\mathbf{g}_i$, while \(J_i^P\) penalizes active passage usage based on passage width and the relative traversal directions of previously planned robots; expired usage records are ignored.
This stage determines the spatial route without assigning times (see Fig.~\ref{fig:single-agent-planning}(a)).

The selected route $\rho_i$ restricts space-time planning to ST-GCS vertices inherited from its constituent spatial regions:
\begin{equation}
\mathcal{V}_{ST}^{\rho_i}
\triangleq
\left\{
v\in\mathcal{V}_{ST}
\,\middle|\,
k(v)\in\rho_i
\right\},
\quad
\mathcal{G}_{ST}^{\rho_i}
\triangleq
\mathcal{G}_{ST}
\left[
\mathcal{V}_{ST}^{\rho_i}
\right].
\label{eq:route_induced_stgcs}
\end{equation}
This retains all temporal subdivisions of the selected regions while excluding vertices outside the route (see Fig.~\ref{fig:single-agent-planning}(b)).
At planning time $\tau$, we introduce the source and goal sets
$
\mathcal{X}_{i}^{s}
=
\left\{
[(\mathbf{p}_i(\tau))^{\top},\tau]^{\top}
\right\},
\ 
\mathcal{X}_{i}^{g}
=
\left\{
[(\mathbf{g}_{i})^{\top},t]^{\top}
\,\middle|\,
t\in[\tau,t_{\max}]
\right\},
$
and connect them to intersecting vertices of $\mathcal{G}_{ST}^{\rho_i}$.
Search proceeds over prefixes of ST vertices $\nu \triangleq (v_1,\ldots,v_M)$. For each prefix, a continuous convex subproblem selects the entry and exit space-time states within the regions, minimizing elapsed time while enforcing inter-region continuity, positive traversal durations, the velocity bound, and the initial state:
\begin{equation}
\begin{aligned}
& \min_{\{\mathbf{z}_{\ell}^{\mathrm{in}},
\mathbf{z}_{\ell}^{\mathrm{out}}\}}
\quad
t_{M}^{\mathrm{out}}-\tau \\
\mathrm{s.t.}\quad
&
\mathbf{z}_{\ell}^{\mathrm{in}},
\mathbf{z}_{\ell}^{\mathrm{out}}
\in
\mathcal{X}_{v_{\ell}},
\quad
\ell=1,\ldots,M,
\\
&
\mathbf{z}_{\ell}^{\mathrm{out}}
=
\mathbf{z}_{\ell+1}^{\mathrm{in}},
\quad
\ell=1,\ldots,M-1,
\\
&
t_{\ell}^{\mathrm{out}}-t_{\ell}^{\mathrm{in}}
\geq
\epsilon,
\quad
\ell=1,\ldots,M,
\\
&
\left\|
\mathbf{p}_{\ell}^{\mathrm{out}}
-
\mathbf{p}_{\ell}^{\mathrm{in}}
\right\|_{2}
\leq
\bar{v}
\left(
t_{\ell}^{\mathrm{out}}
-
t_{\ell}^{\mathrm{in}}
\right),
\quad
\ell=1,\ldots,M,
\\
&
\mathbf{z}_{1}^{\mathrm{in}}
=
[(\mathbf{p}_i(\tau))^{\top},\tau]^{\top},
\end{aligned}
\label{eq:prefix_conditioned_stgcs}
\end{equation}
where
$
\mathbf{z}_{\ell}^{\mathrm{in/out}}
=
[(\mathbf{p}_{\ell}^{\mathrm{in/out}})^{\top},
t_{\ell}^{\mathrm{in/out}}]^{\top}.
$
If the prefix reaches the goal vertex, the terminal constraint
$
\mathbf{p}_{M}^{\mathrm{out}}=\mathbf{g}_{i}
$
is additionally imposed. 
Infeasible prefixes are discarded, while feasible ones are ranked by the optimized duration and the passage-conflict cost introduced below. 
A goal-reaching solution yields the time-parameterized reference path $\pi_i$ (see Fig.~\ref{fig:single-agent-planning}(c)).

\subsubsection{Two-Level Soft Coordination Updates}
\label{subsubsec:coordination_update}

Each reference $\pi_i$ updates two coordination signals: passage-region usage records guide subsequent route selection, while space-time conflict indices penalize temporally overlapping usage.

Let \(\nu_i \triangleq (v_{i,1},\ldots,v_{i,M_i})\) denote the ST-vertex sequence associated with the accepted reference path \(\pi_i\), and define
$
\Lambda_{i,\tilde{k}}
\triangleq
\{\ell\mid k(v_{i,\ell})=\tilde{k}\},
\ 
\tilde{k}\in\mathcal{V}_{P}.
$
For each passage region $\tilde{k}$ with $\Lambda_{i,\tilde{k}}\neq\emptyset$, we update
\begin{equation}
\begin{aligned}
&n_{\tilde{k}}
\leftarrow n_{\tilde{k}}+1,
\ 
I_{i,\tilde{k}}
=
\left[
\min_{\ell\in\Lambda_{i,\tilde{k}}}t_{i,\ell}^{\mathrm{in}},
\,
\max_{\ell\in\Lambda_{i,\tilde{k}}}t_{i,\ell}^{\mathrm{out}}
\right],
\\
&\mathbf{d}_{i,\tilde{k}}
=
\sum_{\ell\in\Lambda_{i,\tilde{k}}}
\left(
\mathbf{p}_{i,\ell}^{\mathrm{out}}
-
\mathbf{p}_{i,\ell}^{\mathrm{in}}
\right),
\end{aligned}
\label{eq:passage_usage_update}
\end{equation}
where \(n_{\tilde{k}}\) counts accepted robots using passage region \(\tilde{k}\), while \(I_{i,\tilde{k}}\) and \(\mathbf{d}_{i,\tilde{k}}\) summarize robot \(i\)'s traversal interval and accumulated displacement. Multiple visits are aggregated using the earliest entry, latest exit, and cumulative displacement, yielding a compact descriptor for \(J_i^P\) in~\eqref{eq:region_route_cost}.

For a traversed passage vertex $v=v_{i,\ell}$ with $k(v)\in\mathcal{V}_P$ and traversal interval $[t_{i,\ell}^{\mathrm{in}},t_{i,\ell}^{\mathrm{out}}]$, we temporally refine $\mathcal{X}_v$ as
\begin{equation}
\begin{aligned}
\mathcal{X}_{v^{-}}
&\triangleq
\mathcal{X}_{v}
\cap
\left\{
\mathbf{z}
\,\middle|\,
t\leq t_{i,\ell}^{\mathrm{in}}
\right\},
\\
\mathcal{X}_{v^{0}}
&\triangleq
\mathcal{X}_{v}
\cap
\left\{
\mathbf{z}
\,\middle|\,
t_{i,\ell}^{\mathrm{in}}
\leq t
\leq
t_{i,\ell}^{\mathrm{out}}
\right\},
\\
\mathcal{X}_{v^{+}}
&\triangleq
\mathcal{X}_{v}
\cap
\left\{
\mathbf{z}
\,\middle|\,
t\geq t_{i,\ell}^{\mathrm{out}}
\right\}.
\end{aligned}
\label{eq:passage_vertex_split}
\end{equation}
Only nonempty portions are retained, and their conflict indices are updated as
$
\kappa_{v^{-}}
=
\kappa_{v^{+}}
=
\kappa_{v},
\ 
\kappa_{v^{0}}
=
\kappa_{v}+1.
$
Repeated updates may further refine existing temporal subdivisions, after which the ST-GCS connectivity is updated according to nonempty set intersections.
For an ST vertex sequence \(\nu\), the search cost combines travel time with a width-weighted passage-conflict penalty:
\begin{equation}
\begin{aligned}
J_i^{ST}(\nu)
\triangleq
J^{T}(\nu)
+
J_i^{C}(\nu),
\ 
J_i^{C}(\nu)
\triangleq
\lambda_C
\sum_{\substack{v\in\nu\\k(v)\in\mathcal{V}_{P}}}
\frac{\kappa_v}{w_{k(v)}},
\end{aligned}
\label{eq:space_time_search_cost}
\end{equation}
where \(J^T(\nu)\) is the optimized travel time from~\eqref{eq:prefix_conditioned_stgcs}, \(w_{k(v)}\) denotes the width of the parent passage region, and \(\lambda_C>0\) weights the conflict penalty. Consequently, narrower and more temporally congested passages incur larger search costs.

The region-level records $(n_{\tilde{k}},I_{i,\tilde{k}},\mathbf{d}_{i,\tilde{k}})$ bias passage-region route selection, whereas $\kappa_v$ biases the subsequent space-time search.
Both provide soft coordination without hard trajectory reservations. The resulting time-parameterized reference $\pi_i$ is passed to the local controller for collision-aware execution.

\subsection{Adaptive MPPI Local Control} \label{sec:mppi}

The local controller receives the global reference $\pi_i$ as time-stamped waypoints $(\mathbf{p}_{i,\ell},t_{i,\ell})$, using them as soft guidance for receding-horizon MPPI control~\cite{williams2016aggressive}. At each control cycle, MPPI samples control sequences around the shifted previous solution, propagates them through the robot dynamics, and executes the first control of the resulting cost-weighted update. To strengthen collision avoidance beyond cost penalties, we adopt the distribution-shaping mechanism of MPPI-ORCA~\cite{dergachev2024model,dergachev2025decentralized}, which maps ORCA velocity constraints into the control space, adjusts the first-step sampling distribution, and rejects samples that violate these constraints. Building on this formulation, we incorporate static-obstacle avoidance and adaptive spatiotemporal reference tracking, as described below.

\subsubsection{Static-Obstacle Avoidance}
Static obstacles are incorporated into both first-step distribution shaping and rollout evaluation. ORCA constraints derived from obstacle boundaries are combined with inter-agent constraints to adjust the first-step sampling distribution and reject inadmissible samples.

For rollout evaluation, a predicted state is classified as colliding if the robot disk enlarged by $\varepsilon_{o}$ intersects an obstacle boundary or robot center lies on the obstacle side of the nearest oriented segment and its orthogonal projection falls within that segment. Each colliding state contributes a fixed penalty, making the trajectory penalty proportional to the number of such states. If the trajectory is collision-free, its obstacle cost is obtained by averaging exponentially decaying proximity penalties over the evaluated states:
\begin{equation}
q_{obs}(\mathbf{x_h})
=
\begin{cases}
w_{o.col.}, &\mathbb{I}_{\mathrm{o}}(\mathbf{x_h})=1 \\ 
\frac{w_{o}}{H}\cdot e^{\,-\eta_{o}\big(d_{o}(\mathbf{x_h})-r-\varepsilon_{o}\big)},& \text{o.w.}
\end{cases},
\end{equation}
\noindent where $\mathbb{I}_o$ is the collision indicator, $d_o$ is the distance to the nearest considered obstacle, \(w_{o.col.}\) and \(w_o\) are the collision and proximity weights, and \(\eta_o>0\) controls the decay rate. Near the final goal, the additional buffer is removed, the proximity weight is reduced, and the evaluation horizon is shortened to facilitate the final approach.

\subsubsection{Spatiotemporal Reference Tracking}
Temporal tracking preserves global passage coordination while the robot remains synchronized with its reference. When execution deviations are substantial, spatial tracking follows the planned route according to traveled distance without enforcing nominal timing.

The reference $\pi_i$ is discretized by linear interpolation of positions and timestamps. 
Omitting robot and sample indices, $k$ denote the current execution time step and $t_\ell$ the timestamp of the reference point closest to $\mathbf{p}_k$. 
Temporal tracking selects $\mathbf{p}_h^{\mathrm{tmp}}$ with the timestamp closest to $k+h$. 
Spatial tracking selects the first point $\mathbf p_h^{\mathrm{spc}}$ whose distance along the reference from the point closest to $\mathbf p_k$ reaches or exceeds the rollout's traveled distance. The running cost switches between these targets according to the schedule discrepancy:
\begin{equation}
q_{ref}(\mathbf{x}_h)
=
\begin{cases}
w_{tmp}\|\mathbf p_h-\mathbf p_h^{\mathrm{tmp}}\|_2,
& | k -t_\ell|<\tau_{del},\\
w_{spc}\|\mathbf p_h-\mathbf p_h^{\mathrm{spc}}\|_2,
& \text{otherwise},
\end{cases}
\end{equation}
\noindent where $w_{tmp}$ and $w_{spc}$ are tracking weights.
Running cost is complemented by a terminal penalty $\phi_{point}(\mathbf x_H)$, that attracts the predicted endpoint toward a reference lookahead point or the reference position for $k+H-1$, with mode-dependent weights $w_{\phi,\mathrm{tmp}}$ and $w_{\phi,\mathrm{spc}}$.
Near the goal, direct goal attraction replaces running tracking. Its weight is reduced when a neighbor is nearby to facilitate yielding, and restored when the robot crosses a larger goal-distance threshold or the neighbor moves away.

\subsubsection{Overall Cost Formulation}
Finally, the trajectory cost combines the proposed reference-tracking and static-obstacle  terms $q_{ref}$, $\phi_{point}$, $q_{obs}$ with the interaction and motion-regularization terms from~\cite{dergachev2025decentralized,dergachev2025corl}:
\begin{equation}
\begin{aligned}
&\begin{split}
q(\mathbf{x_h}) &=  
q_{ca}(\mathbf{x_h})+q_{side}(\mathbf{x_h})+\;\ldots\\
&q_{smooth}(\mathbf{x_h})+q_{ref}(\mathbf{x_h})+q_{obs}(\mathbf{x_h}), \\
\end{split}\\
&\phi(\mathbf{x_{H}}) = \phi_{point}(\mathbf{x_{H}}),\\
\end{aligned}
\label{eq:cost_terms}
\end{equation}
where the term $q_{ca}$ penalizes predicted collisions and insufficient separation using nearest-neighbor and average distances, $q_{side}$ promotes consistent passing, and $q_{smooth}$ discourages abrupt speed reductions and reversals. Each term has a separate weight controlling its relative contribution, which is omitted from \eqref{eq:cost_terms} for clarity.

\subsection{Overall Framework} \label{subsec:overall_framework}

Initially, robots wait at their start positions while the global planner selects passage-aware routes $\rho_i$ and computes route-restricted space-time references $\pi_i$ in priority order. Each accepted reference updates passage-usage records and space-time conflict indices for subsequent planning; execution begins once all references are ready.
During execution, receding-horizon MPPI tracks references while avoiding robots and obstacles, using temporal tracking for coordinated timing and spatial tracking to accommodate deviations without strict schedule adherence.
Meanwhile, the framework monitors execution progress and reference consistency. At time $\tau$, the replanning set $\mathcal{A}^{\mathrm{rp}}$ contains robots unable to reach scheduled segment endpoints under the velocity bound, remaining outside the endpoint tolerance afterward, or persistently exceeding an off-track threshold. For each $i\in\mathcal{A}^{\mathrm{rp}}$, previous coordination records are withdrawn, and the global planner computes a new reference $\pi_i^{\mathrm{new}}$ from $\mathbf{p}_i(\tau)$ to $\mathbf{g}_i$ using the incrementally maintained ST-GCS. The new reference updates coordination records and replaces the local reference, while unaffected robots continue executing their existing references.
\section{Experimental Validation}

\subsection{Simulation Setup}
All simulated robots were modeled as disks of radius $r=0.1\,\mathrm{m}$ with differential-drive kinematics. The linear and angular velocities were bounded by $v\in[-1.0,1.0]\,\mathrm{m/s}$ and $\omega\in[-2.0,2.0]\,\mathrm{rad/s}$, respectively. Simulation and control updates were performed at $\Delta t=0.1\,\mathrm{s}$.
The proposed framework was implemented in Python and C++, with both the global and local planners written in C++ and exposed through Python bindings. The graph of convex sets was constructed using $100{,}000$ random samples. The MPPI controller used $B=1{,}500$ sampled trajectories per update and a prediction horizon of $H=15$ steps.

\subsection{Benchmark}


\subsubsection{Setup}
\begin{figure*}[t]
    \centering
    \includegraphics[width=0.99\linewidth]{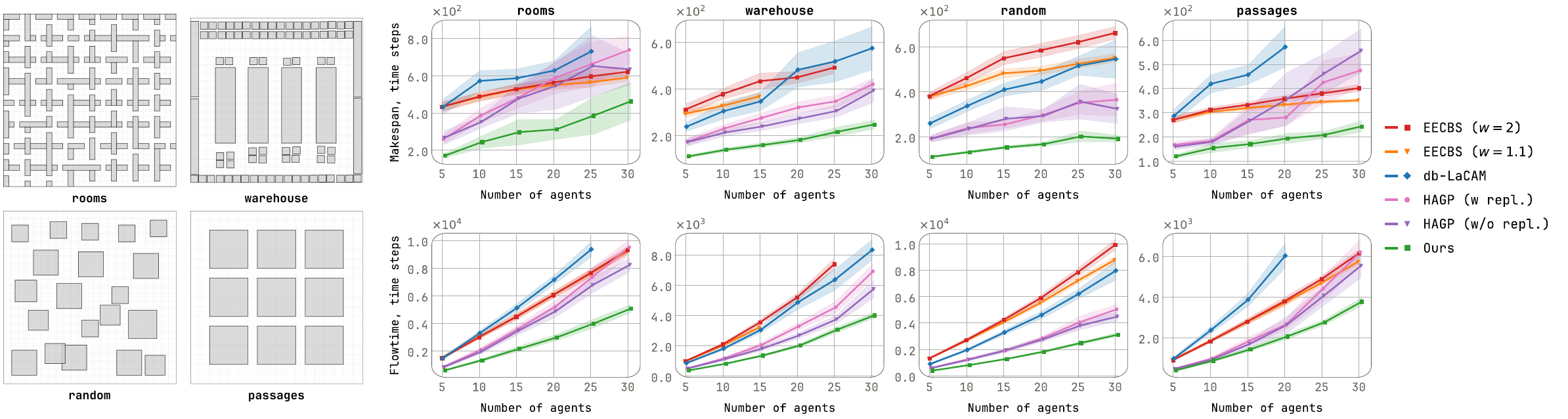}
    \caption{
    Left: environments used in benchmarks, map scales are adjusted for clarity. Right: average makespan (top) and flowtime (bottom) versus team size for the evaluated methods. Lines indicate mean values, and shaded regions show 95\% confidence intervals. Lower values are better.
    }
    \label{fig:maps_metrics}
    \vspace{-0.3cm}
\end{figure*}

We evaluated four maps with alternative routes and varying passage widths (Fig.~\ref{fig:maps_metrics}): \texttt{rooms}, adapted from MovingAI's \texttt{room-32-32-4}~\cite{stern2019multi} ($32\times32$ cells of size $0.45\,\mathrm{m}$); \texttt{warehouse}, scaled from Bench-MR~\cite{heiden2021benchmr} ($9.9\times9.45\,\mathrm{m}$); \texttt{random}, with randomly placed square obstacles ($10.35\times10.35\,\mathrm{m}$); and \texttt{passages}, with nine square obstacles in a $3\times3$ array forming four intersecting $0.45\,\mathrm{m}$-wide corridors ($8.1\times8.1\,\mathrm{m}$). All maps had impassable outer boundaries. For each map, we generated 50 instances of 30 randomly sampled start--goal pairs in free space (restricted to peripheral regions in \texttt{passages}). Each instance was tested with the first $N$ pairs for $N\in\{5,10,15,20,25,30\}$, yielding 1,200 test cases.

We compared four baselines with distinct coordination strategies: \textsc{ORCA-DD}~\cite{snape2010smooth} combines reactive collision avoidance and Dijkstra algorithm on visibility graphs, \textsc{EECBS}~\cite{li2021eecbs} resolves conflicts through discrete MAPF search, \textsc{db-LaCAM}~\cite{moldagalieva2026dblacam} performs kinodynamic multi-robot planning, and \textsc{HAGP}~\cite{dong2025homotopy} combines homotopy-aware global coordination with local control.
\textsc{ORCA-DD} used twice the physical robot radius plus a $0.01\,\mathrm{m}$ margin to compensate for numerical errors and a holonomic speed limit of $0.2\,\mathrm{m/s}$ to respect the angular-speed limit. Paths on visibility graphs were replanned upon a $0.2\,\mathrm{m}$ deviation or backward progress.
\textsc{EECBS} was run with suboptimality factors \(w \in \{1.1, 2.0\}\) on maps discretized into \(0.45\,\mathrm{m}\) cells. Its plans were executed synchronously, with each step comprising a one-cell translation followed by a rotation of up to $90^\circ$ at maximum speed, during which nonrotating robots waited.
For \textsc{db-LaCAM}, planned linear and angular velocity commands $(v,\omega)$ were replayed open-loop, with robot states updated via forward-Euler integration of differential-drive kinematics.
For \textsc{HAGP}, we replaced its MPC controller with our MPPI controller for a common differential-drive local controller and tested variants with and without global replanning.

\begin{table}[t]
\centering
\caption{\textbf{Success rate} across map types. higher values are better; the best values in each column are highlighted in bold.}
\label{tab:sr}
\footnotesize
\setlength{\tabcolsep}{2pt}
\renewcommand{\arraystretch}{1.0}
\begin{tabularx}{\columnwidth}{@{}>{\raggedright\arraybackslash}p{2.3cm}*{4}{Y}@{}}
\toprule
& \texttt{rooms}
& \texttt{warehouse}
& \texttt{random}
& \texttt{passages} \\
\midrule
\textsc{ORCA-DD}
& 20\% 
& 10.3\%  
& 25.7\%  
& 11\%  \\
\textsc{EECBS} ($w=2$)
& \textbf{100\%} 
& 82.3\%
& \textbf{100\%} 
& \textbf{100\%} \\
\textsc{EECBS} ($w=1.1$)
& \textbf{100\%} 
& 49.3\%  
& 96.7\% 
& \textbf{100\%} \\
\textsc{db-LaCAM}
& 65.3\% 
& 78.3\% 
& 84.7\% 
& 60.3\% 
\\
\textsc{HAGP} (w/ repl.)
& 97.3\% 
& 99.3\% 
& \textbf{100\%} 
& 99.3\%  \\
\textsc{HAGP} (w/o repl.)
& 97.7\% 
& \textbf{100\%} 
& \textbf{100\%} 
& \textbf{100\%} \\
\midrule
\textsc{Ours}
& 98.7\% 
& \textbf{100\%} 
& \textbf{100\%} 
& \textbf{100\%}  \\
\bottomrule
\end{tabularx}
\end{table}


All methods were tested on identical instances. A run succeeded if all robots reached and remained within $0.15\,\mathrm{m}$ of their goals without collisions within $2{,}000$ simulation steps. \textsc{EECBS} and \textsc{db-LaCAM} had a $60\,\mathrm{s}$ search budget; all recorded initial-planning and individual replanning calls of the other methods took less than $10\,\mathrm{s}$ (mean $\leq 2\,\mathrm{s}$). We report \textbf{success rate} (fraction of successful runs), \textbf{makespan} (latest robot arrival time), and \textbf{flowtime} (sum of individual arrival times). Arrival is defined as the final entry into the goal region without subsequent departure.

\subsubsection{Success Rate}
Our method achieved near-perfect success rates across all environments, reaching $98.7\%$ on \texttt{rooms} and $100\%$ on the remaining maps, as shown in Table~\ref{tab:sr}.
\textsc{EECBS} reached $100\%$ on \texttt{rooms}, potentially because its explicit conflict resolution better handles goal-occupancy dependencies that challenge our fixed-priority planner. However, its lower success on \texttt{warehouse} may reflect constraints imposed by grid discretization and increased search difficulty; raising $w$ from $1.1$ to $2.0$ improved success from $49.3\%$ to $82.3\%$. \textsc{HAGP} also maintained high success, while \textsc{ORCA-DD} failed on most tasks, consistent with conservative radius inflation and limited global coordination. \textsc{db-LaCAM} outperformed \textsc{ORCA-DD} but was less reliable than the hierarchical methods. Most failures involved incomplete task execution rather than collisions. Rare collisions occurred only for \textsc{db-LaCAM} on \texttt{passages} ($0.67\%$) and \textsc{ORCA-DD} on \texttt{random} ($0.67\%$) and \texttt{passages} ($1\%$), possibly due to numerical instability.

\begin{figure}[t]
    \centering
    \includegraphics[width=\linewidth]{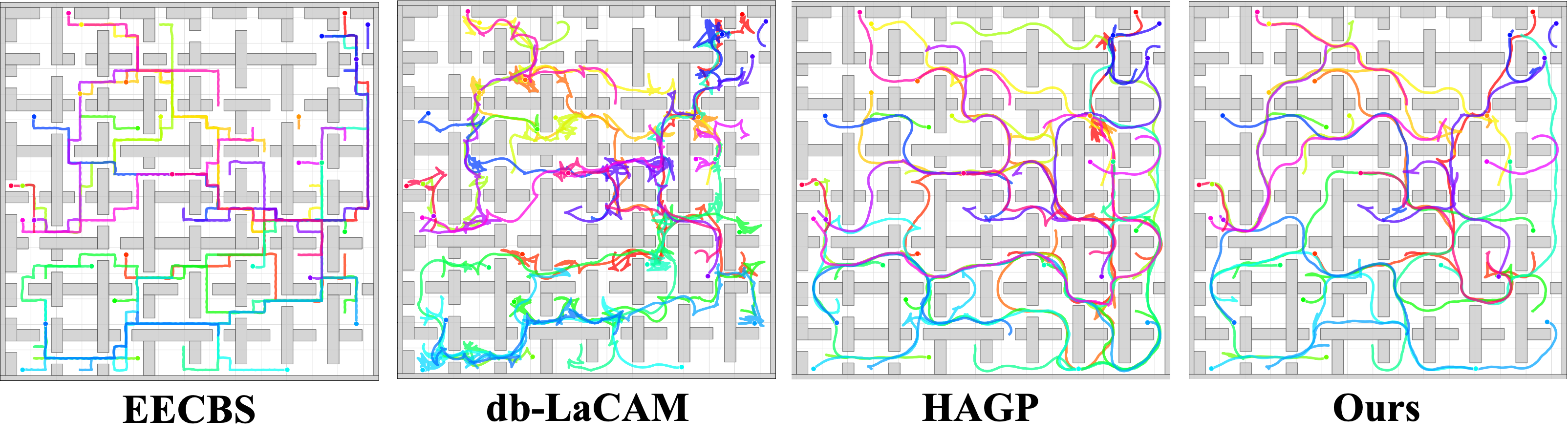}
    \caption{Representative trajectories of 30 robots on \texttt{rooms} for key evaluated algorithms. \textsc{ORCA-DD}, \textsc{HAGP} (w/ repl.) and \textsc{EECBS} ($w=2$) were omitted for clarity. 
    Videos demonstrating robot motion in the numerical experiments for all four maps are provided in the supplementary material.
    }
    \label{fig:trajectories}
    \vspace{-0.1cm}
\end{figure}%
\begin{table}[t]
\centering
\caption{Global planner ablation. Median (IQR) is reported.}
\label{tab:global_planner_ablation}
\begin{threeparttable}
\footnotesize
\setlength{\tabcolsep}{2pt}
\renewcommand{\arraystretch}{1.0}

\begin{tabular}{@{}lccccc@{}}
\toprule
Variant
& GP (s) $\downarrow$
& MS (s) $\downarrow$
& FT (s) $\downarrow$
& $P_{\mathrm{pl}}$ $\downarrow$
& $P_{\mathrm{ex}}$ $\downarrow$ \\
\midrule

\textsc{NoCo}
& \shortstack{12.9 (5.1)}
& \shortstack{44.4 (19.8)}
& \shortstack{878.9 (258.2)}
& \shortstack{534.4 (117.5)}
& \shortstack{748.5 (335.4)} \\

\textsc{RegOnly}
& \shortstack{4.6 (0.8)}
& \shortstack{41.8 (28.0)}
& \shortstack{801.2 (372.3)}
& \shortstack{442.3 (381.2)}
& \shortstack{631.7 (469.6)} \\

\textsc{STOnly}
& \shortstack{9.1 (1.9)}
& \shortstack{37.5 (11.6)}
& \shortstack{829.3 (176.7)}
& \shortstack{\textbf{301.2} (126.5)}
& \shortstack{616.0 (219.4)} \\

\textsc{FullG}
& \shortstack{51.8 (5.7)}
& \shortstack{39.9 (18.0)}
& \shortstack{763.3 (262.7)}
& \shortstack{408.1 (218.3)}
& \shortstack{506.6 (597.0)} \\

\midrule

\textsc{Ours}
& \shortstack{\textbf{4.3} (0.6)}
& \shortstack{\textbf{35.4} (17.3)}
& \shortstack{\textbf{732.2} (243.3)}
& \shortstack{362.8 (154.4)}
& \shortstack{\textbf{480.8} (499.5)} \\

\bottomrule
\end{tabular}

\begin{tablenotes}[flushleft]
\scriptsize
\item
GP: global planning time;
MS: makespan;
FT: flowtime;
$P_{\mathrm{pl}}$/$P_{\mathrm{ex}}$:
planned/executed passage pressure defined by \eqref{eq:passage_pressure}
(robot-pair$\cdot$s).
Bold indicates the lowest median.

\item
\textsc{RegOnly}: omit $J_i^C(\nu)$
in~\eqref{eq:space_time_search_cost};
\textsc{STOnly}: omit $J_i^P(\rho)$
in~\eqref{eq:region_route_cost};
\textsc{NoCo}: both $J_i^P$ and $J_i^C$ disabled;
\textsc{FullG}: direct full-graph search
with ST coordination only;
\textsc{Ours}: both coordination terms enabled with route-restricted ST-GCS search.
\end{tablenotes}

\end{threeparttable}
\end{table}

\subsubsection{Execution Efficiency}

Our method achieved the lowest mean makespan and flowtime across environments and team sizes, as shown in Fig.~\ref{fig:maps_metrics}. At each point, we included only methods with success rates of at least $50\%$ and instances completed by all included methods; \textsc{ORCA-DD} qualified at too few points to be shown.
On \texttt{rooms} with 30 robots, the best competing makespan and flowtime were approximately $1.3$ and $1.6$ times ours, respectively. 
\textsc{HAGP} was generally the closest competitor, although \textsc{EECBS} achieved lower makespan than \textsc{HAGP} for larger teams on \texttt{rooms} and \texttt{passages}, potentially benefiting from explicit conflict resolution.
Global replanning provided limited improvements for \textsc{HAGP}. In the representative 30-robot \texttt{rooms} example (Fig.~\ref{fig:trajectories}), all methods succeeded, but our method exhibited the least oscillation near narrow passages among continuous-space methods, while \textsc{EECBS} required longer execution and utilized free space less effectively.

\subsection{Ablation Studies}

\begin{table}[t]
\centering
\caption{Robot-count scalability. Median (IQR) is reported.}
\label{tab:global_planner_scalability}
\begin{threeparttable}
\footnotesize
\setlength{\tabcolsep}{8pt}
\renewcommand{\arraystretch}{1.0}

\begin{tabular}{@{}ccccc@{}}
\toprule
Robots
& GP (s)
& Exact Calls
& MS (s)
& FT/Robot (s) \\
\midrule

8  & 0.9 (0.1) & 170.0 (21.0)
   & 17.7 (3.6) & 13.8 (1.9) \\

16 & 2.0 (0.2) & 384.0 (49.2)
   & 21.9 (7.5) & 15.4 (3.5) \\

24 & 3.2 (0.4) & 612.0 (73.2)
   & 26.8 (5.6) & 18.6 (2.8) \\

32 & 4.3 (0.6) & 827.0 (113.8)
   & 35.4 (17.3) & 22.9 (7.6) \\

\bottomrule
\end{tabular}

\begin{tablenotes}[flushleft]
\scriptsize
\item
GP: global planning time;
Exact Calls: total exact solver calls;
MS: makespan;
FT/Robot: flowtime divided by the robot count.
\end{tablenotes}

\end{threeparttable}
\end{table}

\subsubsection{Global Planner Analyses} \label{sec:global-planner-sim}
We evaluate five planner variants on 20 randomly generated $10.35\,\mathrm{m}$ square maps, each with 12 rectangular obstacles. In each scenario, 32 robots start uniformly along the inner perimeter and perform reciprocal swaps. All variants execute a single initial plan without replanning using the same MPPI controller (Sec.~\ref{sec:mppi}).
To quantify bottleneck congestion, we define passage pressure as
\begin{equation}
P_{m}
\triangleq
\sum_{p\in\mathcal{P}}
\int_{0}^{T_m}
\binom{n_p^{m}(t)}{2}\,dt,
\quad m\in\{\mathrm{pl},\mathrm{ex}\},
\label{eq:passage_pressure}
\end{equation}
where $\mathcal{P}$ is the passage set, $T_m$ is the evaluation horizon, and $n_p^m(t)$ counts robots simultaneously occupying passage $p$ in planned ($\mathrm{pl}$) or executed ($\mathrm{ex}$) trajectories. The integral of $\binom{n_p^m(t)}{2}$ measures cumulative robot-pair co-occupancy time (robot-pair$\cdot$s).
Table~\ref{tab:global_planner_ablation} reports medians and interquartile ranges (IQRs) across 20 scenarios.
Compared with \textsc{NoCo} (no coordination), \textsc{Ours} reduces median GP time and makespan by $66.7\%$ and $20.3\%$, respectively.
Relative to \textsc{RegOnly} (region-level only), it reduces makespan and planned pressure by $15.3\%$ and $18.0\%$; relative to \textsc{STOnly} (space-time only), it reduces GP time and executed pressure by $52.7\%$ and $21.9\%$.
Compared with \textsc{FullG} (full-graph search with space-time coordination), it reduces GP time by $91.7\%$.
These results demonstrate the benefits of two-level soft coordination and route-restricted planning for efficient closed-loop navigation.

\begin{table}[t]
\centering
\caption{Replanning strategies comparison. Median (IQR) is reported.}
\label{tab:replanning_comparison}
\begin{threeparttable}
\footnotesize
\setlength{\tabcolsep}{4pt}
\renewcommand{\arraystretch}{1.0}

\begin{tabular}{@{}lcccc@{}}
\toprule
Method
& MS (s) $\downarrow$
& FT (s) $\downarrow$
& $P_{\mathrm{ex}}$ $\downarrow$
& $T_{\mathrm{rep}}$ (ms) $\downarrow$ \\
\midrule

\textsc{Fixed}
& \shortstack{\textbf{35.4} (17.3)}
& \shortstack{\textbf{732.2} (243.3)}
& \shortstack{480.8 (499.5)}
& -- \\

\textsc{Periodic}
& \shortstack{37.9 (14.0)}
& \shortstack{756.0 (189.7)}
& \shortstack{506.2 (299.0)}
& \shortstack{2217.3 (485.5)} \\

\textsc{Selective}
& \shortstack{36.2 (9.5)}
& \shortstack{741.0 (144.8)}
& \shortstack{\textbf{451.4} (215.7)}
& \shortstack{\textbf{156.8} (28.1)} \\

\bottomrule
\end{tabular}

\begin{tablenotes}[flushleft]
\scriptsize
\item
MS: makespan;
FT: flowtime;
$P_{\mathrm{ex}}$: executed passage pressure
(robot-pair$\cdot$s);
$T_{\mathrm{rep}}$: mean time per
global replanning call within each run.
\item
\textsc{Fixed}: no replanning;
\textsc{Periodic}: replanning all robots every 5 seconds;
\textsc{Selective}: proposed replanning mechanism.
Bold indicates the lowest median.
\end{tablenotes}

\end{threeparttable}
\end{table}

\begin{figure}[b!]
    \centering
    \includegraphics[width=\linewidth]{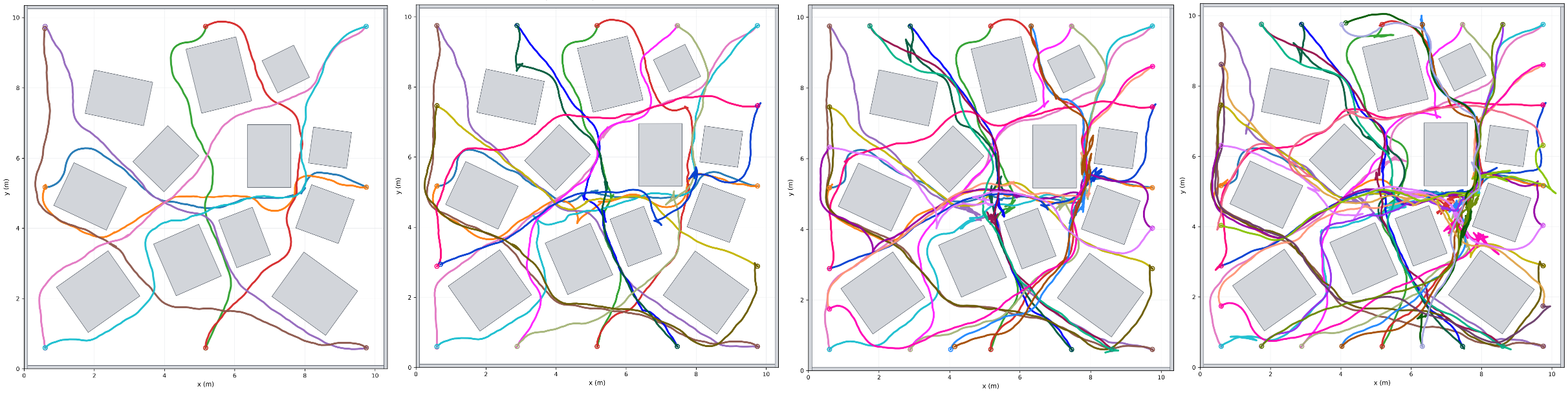}
    \caption{Robot-count scalability in a representative randomly generated scenario. Executed trajectories of \textsc{Ours} with 8, 16, 24, and 32 robots (left to right), using nested agent subsets with reciprocal perimeter start--goal swaps.}
    \label{fig:scale}
    \vspace{-0.2cm}
\end{figure}

We further assess scalability from 8 to 32 robots using nested subsets of the same 20 scenarios (Fig.~\ref{fig:scale}). Table~\ref{tab:global_planner_scalability} shows that median GP time and exact solver calls increase by factors of approximately $4.8$ and $4.9$, respectively, while makespan doubles and per-robot flowtime rises by only $66\%$. All 80 runs succeed, although makespan IQR grows nearly fivefold, indicating greater variability at higher densities.

\subsubsection{Local Planner Analyses} \label{sec:local-planner-sim}

\begin{figure}[h!]
\centering
\begin{minipage}[c]{0.16\columnwidth}
    \centering
    \includegraphics[width=\linewidth]{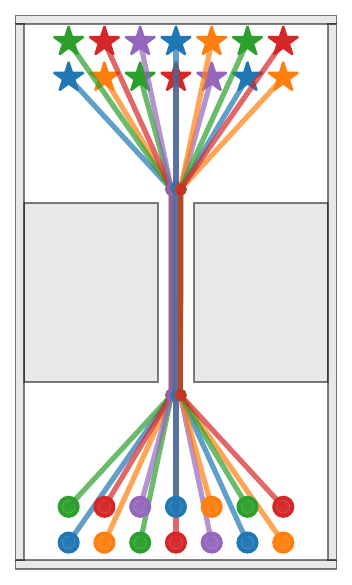}
\end{minipage}
\hfill
\begin{minipage}[c]{0.82\columnwidth}
    \centering
    \small
    \setlength{\tabcolsep}{6pt}
    \renewcommand{\arraystretch}{1.0}
    \begin{tabular}{@{}lcc@{}}
    \toprule
     & MS (steps) $\downarrow$ & FT (steps) $\downarrow$ \\
    \midrule
    Spatial MPPI
    & $201.4 \pm 62.3$
    & $1774.4 \pm 344.9$ \\
    Temporal MPPI
    & $569.1 \pm 113.7$
    & $6307.7 \pm 1269.7$ \\
    \midrule
    Adaptive MPPI
    & $\mathbf{190.8} \pm 49.3$
    & $\mathbf{1727.9} \pm 304.2$ \\
    \bottomrule
    \end{tabular}
\end{minipage}
\caption{Local-controller ablation.
Left: the bottleneck environment.
Right: makespan (MS) and flowtime (FT) of the local-controller variants, reported as mean $\pm$ standard deviation. Bold indicates the lowest mean.}
\label{fig:local_ablation}
\end{figure}

We evaluate local reference tracking in a bottleneck scenario
where 14 robots must traverse a narrow corridor
(Fig.~\ref{fig:local_ablation}). Efficient execution requires
regulating corridor entry to avoid congestion without
unnecessary waiting. We compare spatial-only, temporal-only,
and adaptive tracking over 500 runs with fixed start--goal
assignments and different MPPI random seeds.

\begin{figure*}[th!]
    \centering
    \includegraphics[width=0.99\textwidth]{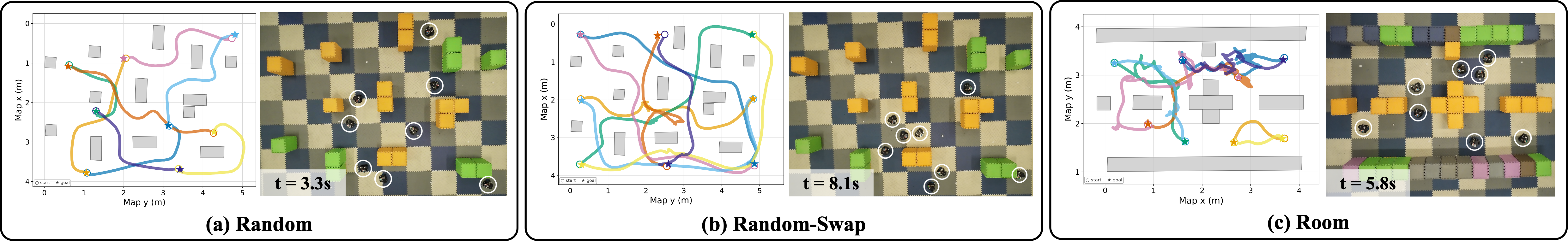}
    \caption{
    Representative hardware experiments with eight differential-drive robots in Random, Random-Swap, and Room (left to right) scenarios. Each panel pairs recorded trajectories with a top-view snapshot. Videos demonstrating robot motion are provided in the supplementary material.
    }
    \label{fig:hardware_experiments}
    \vspace{-0.3cm}
\end{figure*}
Adaptive tracking achieves lower mean makespan and flowtime
than both alternatives, with lower standard deviations
indicating more consistent performance. The improvement over
temporal-only tracking is pronounced because the global planner assigns conservative timestamps in this
scenario. Delayed robots still pursue targets dictated by
the schedule and may get stuck at corners.
Robots entering ahead of schedule may instead attempt to
retreat, obstructing those proceeding on schedule.
Spatial fallback allows adaptive tracking to accommodate
these deviations and maintain progress.

\subsubsection{Selective Replanning Analysis}
\label{sec:replanning-sim}

Using the same 20 scenarios and 32-robot tasks as
Sec.~\ref{sec:global-planner-sim}, we compare fixed
global references (\textsc{Fixed}), periodic
replanning of all robots every 5\,s
(\textsc{Periodic}), and our event-triggered,
agent-selective replanning (\textsc{Selective})
under the same MPPI controller.
As shown in Table~\ref{tab:replanning_comparison},
compared with \textsc{Fixed}, \textsc{Selective}
reduces median executed passage pressure by $6.1\%$,
while increasing median makespan and flowtime by
only $2.3\%$ and $1.2\%$, respectively.
Meanwhile, the IQRs of makespan, flowtime, and
executed pressure decrease by $45.1\%$, $40.5\%$,
and $56.8\%$, respectively, indicating reduced
inter-scenario performance variability.
Compared with \textsc{Periodic}, \textsc{Selective}
reduces median makespan, flowtime, and executed
passage pressure by $4.5\%$, $2.0\%$, and
$10.8\%$, respectively.
Its mean per-call replanning latency is also
$92.9\%$ lower, highlighting its ability to perform
lightweight, targeted online replanning by updating
only the triggered robots rather than all robots.

\subsection{Hardware Experiments}


\subsubsection{Setup}
Hardware experiments used eight differential-drive robots (radius $0.1\,\mathrm{m}$) in an indoor environment with OptiTrack localization. A desktop (Intel Core i7-13700K, 3.4 GHz; 32 GB RAM) performed all planning and control computations and transmitted commands to the robots. Velocity limits were $v\in[-0.6,0.6]\,\mathrm{m/s}$ and $\omega\in[-4.0,4.0]\,\mathrm{rad/s}$.
We tested two environments: Random ($4\times5\,\mathrm{m}$, 13 randomly placed obstacles) and Room ($4\times3\,\mathrm{m}$, nine obstacles forming four rooms connected by five doorways). Table~\ref{tab:hardware_results} summarizes five random start--goal configurations in Random, ten in Room, and two repetitions of a fixed Random-Swap configuration, where robots were evenly distributed along the boundary and exchanged positions with their counterparts (Fig.~\ref{fig:hardware_experiments}). All 17 trials succeeded without collisions.

\begin{table}[t]
    \centering
    \caption{Hardware experiment results. Mean is reported}
    \label{tab:hardware_results}
    \begin{threeparttable}
    \footnotesize
    \setlength{\tabcolsep}{3pt}
    \renewcommand{\arraystretch}{1.0}
    \begin{tabular}{@{}lccccccc@{}}
        \toprule
        Scenario
        & Succ.
        & MS (s)
        & Len (m)
        & R
        & GP (s)
        & $T_{\mathrm{rep}}$ (ms)
        & $T_{\mathrm{ctrl}}$ (ms) \\
        \midrule
        Random
        & 5/5 & 16.83 & 26.63 & 13.4
        & 0.49 & 90.23 & 68.09 \\

        Rand-Swap
        & 2/2 & 21.01 & 52.04 & 7.5
        & 0.97 & 258.53 & 69.89 \\

        Room
        & 10/10 & 23.86 & 28.56 & 27.3
        & 0.40 & 79.05 & 65.11 \\
        \midrule
        Overall
        & 17/17 & 21.46 & 30.75 & 20.9
        & 0.49 & 88.75 & 66.45 \\
        \bottomrule
    \end{tabular}
    \begin{tablenotes}[flushleft]
        \scriptsize
        \item Succ.: successful/total trials.
        MS: makespan; Len: total trajectory length of all robots;
        R: number of replanning requests.
        GP: global planning time.
        $T_{\mathrm{rep}}$: mean time per online
        replanning call.
        $T_{\mathrm{ctrl}}$: mean time per control cycle.
    \end{tablenotes}
    \end{threeparttable}
\end{table}

\subsubsection{Results}
The Room experiments had the largest mean makespan ($23.86\,\mathrm{s}$) and most frequent global replanning ($27.3$ requests per trial), reflecting the coordination challenges posed by constrained doorways. Random-Swap produced the longest total trajectories ($52.04\,\mathrm{m}$), reflecting the coupled position-exchange task. Across all trials, initial global planning for eight robots averaged $0.49\,\mathrm{s}$, while online replanning required $88.75\,\mathrm{ms}$ per request and the online local control computation averaged $66.45\,\mathrm{ms}$ per cycle.

\section{Conclusion}
We presented a hierarchical multi-robot navigation framework combining passage-focused soft global coordination with adaptive MPPI local control. Passage-aware route selection and continuous space-time planning guide bottleneck traversal through soft conflict penalties, while adaptive reference tracking and selective replanning accommodate execution deviations without rigid trajectory reservations. 
Across 1,200 simulation instances, our method achieved near-perfect success rates and the lowest mean makespan and flowtime on common successful instances. All 17 hardware trials with eight robots were completed without collisions.

While these results demonstrate the framework's effectiveness, several trade-offs remain: fixed priorities may cause unfavorable interactions at narrow passages, two-stage planning may miss more efficient detours, and adaptive tracking permits substantial schedule deviations. Addressing these limitations represents a promising direction for future work.

\bibliographystyle{IEEEtran}
\bibliography{IEEEabrv,ref}

\end{document}